\documentclass[conference]{IEEEtran}
\IEEEoverridecommandlockouts

\usepackage{cite}
\usepackage{amsmath,amssymb,amsfonts}
\usepackage{algorithmic}
\usepackage{graphicx}
\usepackage{textcomp}
\usepackage{xcolor}
\usepackage{subfigure}

\def\BibTeX{{\rm B\kern-.05em{\sc i\kern-.025em b}\kern-.08em
    T\kern-.1667em\lower.7ex\hbox{E}\kern-.125emX}}
\begin{document}

\title{Automated Extraction of Spatio-Semantic Graphs for Identifying  Cognitive Impairment \\
}

\author{\IEEEauthorblockN{Si-Ioi Ng$^{1}$, Pranav S. Ambadi$^{1}$, Kimberly D. Mueller$^{2}$, Julie Liss$^{1}$, Visar Berisha$^{1}$}
\IEEEauthorblockA{\textit{$^{1}$Arizona State University, USA} \\ 
\textit{$^{2}$University of Wisconsin-Madison, USA}} 
$^{1}$\{siioing, pambadi, JULIE.LISS, visar\}@asu.edu, $^{2}$kdmueller@wisc.edu}

\maketitle

\begin{abstract}
Existing methods for analyzing linguistic content from picture descriptions for assessment of cognitive-linguistic impairment often overlook the participant's visual narrative path, which typically requires eye tracking to assess. Spatio-semantic graphs are a useful tool for analyzing this narrative path from transcripts alone, however they are limited by the need for manual tagging of content information units (CIUs). In this paper, we propose an automated approach for estimation of spatio-semantic graphs (via automated extraction of CIUs) from the Cookie Theft picture commonly used in cognitive-linguistic analyses. The method enables the automatic characterization of the visual semantic path during picture description. Experiments demonstrate that the automatic spatio-semantic graphs effectively differentiate between cognitively impaired and unimpaired speakers. Statistical analyses reveal that the features derived by the automated method produce comparable results to the manual method, with even greater group differences between clinical groups of interest. These results highlight the potential of the automated approach for extracting spatio-semantic features in developing clinical speech models for  cognitive impairment assessment.

\end{abstract}


\section{Introduction}
\label{sec:intro}
Dementia is a neurodegenerative disease that leads to decline in multiple cognitive domains, affecting an individual's social or occupational function \cite{arvanitakis2019diagnosis}. 
Traditional methods for characterising cognitive impairment involve reviewing an individual's history, collecting data from patient and caregiver questionnaires, conducting neuropathological and neuropsychological tests, etc. 
The growing prevalence of dementia requires automated assessment tools that enable early diagnosis and longitudinal monitoring of cognitive impairment. 
A patient's speech is a prominent biomarker since the speech production process can reflect problems in cognition and perception \cite{berisha2024responsible}. 
Speech analytic systems designed for cognitive impairment are typically based on neuropsychological tests, which involve recalling the keywords in verbal memory tests, or describing the objects and actions in picture description tasks \cite{kertesz1986dissolution, mueller2021amyloid}. When the elicited speech is recorded and transcribed, automated speech and language analytics can be used for early detection of impairment. 

Among the most common ways to elicit speech from patients is the Cookie Theft picture description task from the Boston Diagnostic Aphasia Examination  \cite{goodglass2001bdae}. The task is straight-forward to  administer and can magnify changes occurring in brain regions implicated by cognitive impairment. 
Explained in \cite{cummings2019describing}, 
the diverse features derived from the task, such as semantic categories, referential cohesion, and 
mental state language, etc., are particularly suited to detecting cognitive impairments. Previous studies have developed approaches to detect cognitive impairment using speech processing, natural language processing \cite{luz20_interspeech, luz21_interspeech}. 
Acoustic features such as paralinguistic feature sets, MFCCs, neural network embeddings of speech; and language features such as TF-IDF, text embeddings extracted from large language model, are applied to model training and clinical label prediction \cite{balagopalan20_interspeech, martinc20_interspeech, pappagari20_interspeech, edwards20_interspeech}. 

These existing features are often highly abstract and do not link to clinical constructs of cognitive impairment. As a result, speech analytic systems developed on these features cannot easily be adopted in real-world clinical settings or provide useful feedback to clinicians. 
An alternative approach could be utilizing measures used by clinicians in clinical assessments. Specific to the Cookie Theft task, clinicians  utilize transcriptions of participant descriptions to determine features such as content information units (CIUs), pauses, and repetitions as well as acoustic measures derived from the recorded speech, such as speech rate and voice quality \cite{mueller_connected_2016,mueller_connected_2018}. Less accessible to clinicians, however, is the visual-path analysis of a participant's description. Picture description engages not only the cortical pathways responsible for semantic retrieval and production, but also pathways connecting the parietal lobe with visual processing circuits. Growing evidence indicates that the parietal lobe, implicated in visuospatial tasks, is among the earliest regions to exhibit neurodegenerative changes \cite{jacobs_parietal_2012,salimi_can_2018}. 
Such impairments can manifest in a patient's description of a complex visual scene. Tools that capture the visual relationship between objects in the picture can help clinicians better understand the underlying neuropathology behind cognitive deficits. 
One such tool was recently introduced in the form of spatio-semantic graphs \cite{ambadi2021spatio}. The authors propose a graph-theoretic representation that aims to encode the sequential listing of CIUs and their relative spatial position in the Cookie Theft picture. Results showed significant differences in features extracted from spatio-semantic graphs between cognitively impaired and cognitively unimpaired speakers, reflecting measurable deficits in visuospatial, attentional, and organizational abilities in cognitively impaired speakers. 
A limitation of integrating this tool into clinical speech analytics pipeline is its dependence on manually annotated CIUs to obtain accurate spatio-semantic graphs. It necessitates automating CIU annotation to efficiently extract the features from spatio-semantic graphs. Given the Cookie Theft picture's usage in the largest Alzheimer's disease and dementia cohort studies to date, the availability of these automatically extracted spatio-semantic features will serve as a great aid to clinicians in automatically assessing cognitive impairment \cite{becker1994natural, johnson2018wisconsin, petersen2010alzheimer, liss2024operationalizing}.  

In this study, we automatically extract CIUs from a speaker's Cookie Theft description. 
This is done by automatic mapping of transcripts to CIUs via a word list for each CIU. We describe the word list design and compare the CIUs extracted manually and automatically from the same speech transcription. We then use the resulting CIUs to generate a graph-theoretic structure that represents the speaker's visual semantic path as they describe the Cookie Theft picture.
To evaluate the efficacy of the approach, we visualize and compare the spatio-semantic graphs extracted from a cognitively unimpaired and a cognitively impaired speaker. Furthermore, we compare the statistics of each spatio-semantic features manually-extracted to those automatically-extracted. 

\section{Speech Datasets}
This study utilizes speech data from the Wisconsin Registry for Alzheimer’s Prevention (WRAP) dataset \cite{johnson2018wisconsin} and the DementiaBank (DB) Pitt Corpus \cite{lanzi2023dementiabank}. The two datasets contain 1,058 and 291 speech recordings, respectively. Both speech datasets utilize the Cookie Theft picture description as the speech elicitation task. 
WRAP is a longitudinal, observational cohort  of individuals in midlife, with an emphasis on those with a parental history of Alzheimer’s disease.
Participants attend study visits every two years to provide detailed health and lifestyle data, and undergo comprehensive neuropsychological testing. During each visit, participants are diagnosed into one of the following categories: cognitively unimpaired-stable, cognitively unimpaired-declining, mild cognitive impairment (MCI), or dementia. 
The Pitt Corpus from DB consists of speech recordings collected as part of a large-scale study conducted by the Alzheimer and Related Dementia Study at the University of Pittsburgh School of Medicine. At the time of the speech recording, each participant was assigned a clinical label of cognitively unimpaired, MCI, or dementia.

All speech data were transcribed by trained listeners using the Codes for the Human Analysis of Transcripts (CHAT) format \cite{macwhinney2014childes}. During data pre-processing, to maintain a one-to-one correspondence between speakers and data, only speech data from a patient's first visit was used. 
The cognitively unimpaired data from both DB and WRAP were combined to form the Cognitively Unimpaired group, while the MCI and dementia data were merged into the Cognitively Impaired group. 
Ultimately, the analysis included 1,089 recordings in the Cognitively Unimpaired group and 219 recordings in the Cognitively Impaired group.

\begin{table*}[htbp!]
    \caption{Dictionary for automatic extraction of CIUs}
    \centering
    \resizebox{0.95\linewidth}{!}{
    \begin{tabular}{l|l|l}
         \textbf{Item No.} & \textbf{Content Information Unit} & \textbf{Words related}  \\ \hline 
         1&  Boy & adolescent, boy, brother, child, guy, he, himself, his, kid, male, man, shirt, shoe, sock, son
 \\
         2&  Girl &  child, daughter, dress, girl, kid, sister, skirt\\
         3&  Woman & adult, apron, domestic, dress, fifties, heels, high-heels, homemaker, housewife, lady, ladys, mama, mom, mother, \\ 
         & & nineteen-fifties, wife, woman
 \\
         4&  Kitchen &  counter, culinary, home, interior, kitchen\\
         5&  
         Outside &  background, backyard, bush, day, exterior, flower, garage, garden, grass, grow, lawn, leaves, \\
 & & outdoor, outside, path, pathway, scenery, shrubbery, shrub, sidewalk, snow, spring, summer, sunny, tree, walkway, yard \\
         6&  Cookie & chocolatechip, cookie, pastry, snack \\
         7&  Jar & canister, container, holder, jar, lid, vessel \\
 8&Stool & bench, chair, footstool, furniture, ladder, perch, step, stepladder, stepstool, stool, three, threelegged
 \\
 9&Sink & basin, drain, faucet, sink\\
 10&Plate & dish, plate, saucer\\
 11&Dishcloth  &  cloth, dishcloth, dishrag, dishtowel, handtowel, napkin, rag, textile, towel, sponge \\
 12&Water & deluge, flood, floor, flow, inundation, liquid, moisture, puddle, torrent, water\\
 13&Window & casement, glass, pane, window\\
 14&Cupboard & cabinet, cupboard, door, handle, shelf, storage\\
 15&Dishes & bowl, cup, dish, plate\\
 16&Curtains &curtain, drape, dressing, fabric, hang, textile, tie, tieback, wave, wind, window\\
 17&Boy taking/stealing & acquire, climb, extract, grab, raid, rob, secure, sneak, snitch, steal, snatch, take, try \\
 18&Boy or stool falling & backwards, balance, collapse, crash, fall, hurt, overturn, tilt, tip, tipping, topple, unstable \\
 19&Woman drying/washing plates/dishes &  clean, dry, rinse, scrub, wash, washing, wipe \\
 20&Water overflowing/spilling & deluge, drip, faucet, flood, flow, inundation, overfill, overflow, overflowing, overrun, pour, run, spill, spilling, splash, torrent
 \\
 21&Action performed by the girl & ask, finger, gesture, giggle, laugh, lip, motion, mouth, point, quiet, reach, request, say, shout, shh, signal, speak, tell, warn \\
 22&Woman unconcerned by overflowing & attention, aware, care, concern, daydream, daze, disregard, distract, ignorant, ignore,  \\
 & &neglectful, nonchalant, notice, oblivious, realize, stand, unaware, unconcerned \\
 23& Woman indifferent to the children &apathetic, attention, aware, back, behind, clue, disregard, distract, doesn't see, focus, ignorant, ignore, indifferent, \\
 & & nonchalant, notice, oblivious, turn, unaware, unconcern, unconcerned
\\
    \end{tabular}
    }
    \label{tab:CIU}
\end{table*}

\section{The Cookie Theft Picture and Content Information Units}
\begin{figure}[t!]
  \setlength\belowcaptionskip{-0.8\baselineskip}
  \centering
  \includegraphics[width=0.80\linewidth]{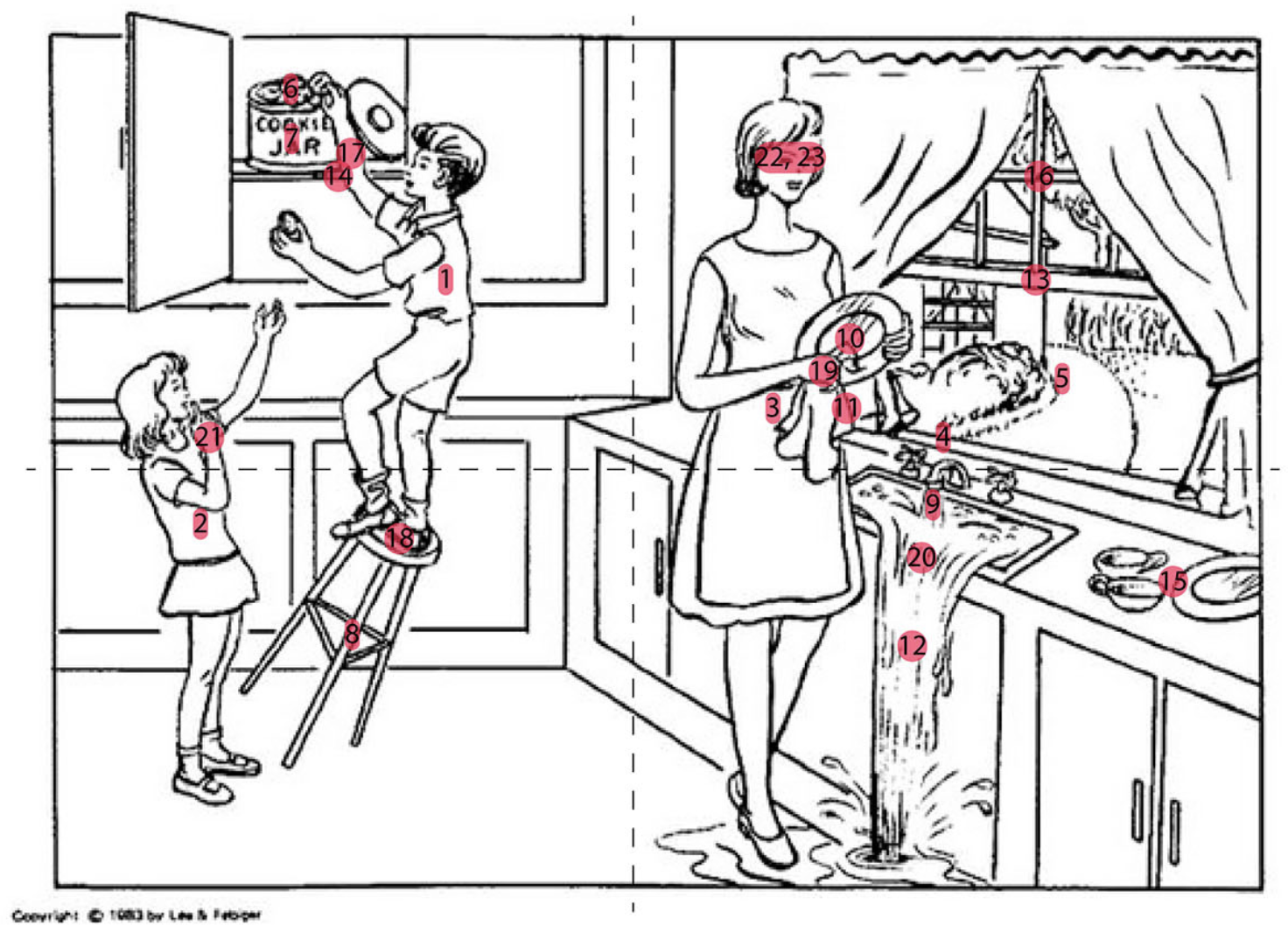}
  \caption{The Cookie Theft picture and CIUs (marked in red) \cite{goodglass2001bdae, ambadi2021spatio}.} 
  \label{fig:cookie-theft}
\end{figure}

Figure \ref{fig:cookie-theft} illustrates the Cookie Theft picture and its 23 CIUs.  The picture depicts a mother drying dishes, unaware of the overflowing sink. Meanwhile, her children try to take cookies: the boy climbs a wobbly stool to reach the jar, while the girl waits with an outstretched hand. Typically, the CIUs are manually tagged from the transcribed speech by trained listeners. 
The list of CIUs and their corresponding words are presented in Table \ref{tab:CIU}. These words were selected by examining the word frequency histogram across the datasets. 
The word selection rules differ from  \cite{mueller2016connected}, as we include more relevant words to capture the visual path during the picture description task, but the dictionary construction remains flexible and can be tailored to the study’s objectives.  

Table \ref{tab:CIU_sequence_example} compares the CIUs sequences extracted manually and automatically (using Table \ref{tab:CIU}) from the same speech example. 
 Although the two sequences are slightly misaligned with each other (as seen in the CIUs highlighted in blue and red), the automated approach encodes CIUs that were skipped by human annotators  (as highlighted in green), which provides a more fine-grained description regarding the visuospatial path of speakers when performing the picture description task. 

\begin{table}
    \caption{Comparison of manually and automatically extracted CIU sequences. Colors indicate discrepancies: Red (automatic annotation errors), Green (human annotator omissions), Blue (automatic annotation misses).}
    \centering
        \resizebox{\linewidth}{!}{
\begin{tabular}{l|l}
\textbf{\begin{tabular}[c]{@{}l@{}}Original\\ Transcript\end{tabular}}                     & \begin{tabular}[c]{@{}l@{}}I see mom's doing the dishes. She has might might be deep in thought, \\ and that’s why the sink is overflowing onto the floor. \\ The kids are up in the cookie jar. She doesn’t seem to be noticing them. \\ Little kid's gonna fall off the chair from getting the cookies. \\ Seems like a nice day outside, and the little girl is kind of telling her brother, \\ “Shh, don’t tell anybody.”\end{tabular}                                                                                                            \\ \hline
\textbf{\begin{tabular}[c]{@{}l@{}}Manually \\ annotated\\ CIU sequence\end{tabular}}      & \begin{tabular}[c]{@{}l@{}}woman --\textgreater dishes --\textgreater \textcolor{blue}{drying/washing plates} --\textgreater sink --\textgreater water overflowing --\textgreater \\ jar --\textgreater woman indifferent to children --\textgreater boy/stool falling \\ --\textgreater cookie --\textgreater outside --\textgreater action by girl --\textgreater girl --\textgreater boy\end{tabular}                                                                                                                                                              \\ \hline
\textbf{\begin{tabular}[c]{@{}l@{}}Automatically \\ annotated\\ CIU sequence\end{tabular}} & \begin{tabular}[c]{@{}l@{}}woman --\textgreater dishes --\textgreater sink --\textgreater water overflowing --\textgreater \textcolor{green}{water} --\textgreater \textcolor{green}{boy} --\textgreater \textcolor{green}{girl} --\textgreater \\ \textcolor{green}{cookie} --\textgreater  jar --\textgreater woman indifferent to children \\ --\textgreater \textcolor{red}{woman unconcerning overflow} --\textgreater   boy --\textgreater \textcolor{red}{girl} --\textgreater boy/stool falling --\textgreater \textcolor{green}{stool} --\textgreater \\ \textcolor{green}{stealing cookie} --\textgreater  cookie --\textgreater  outside --\textgreater girl --\textgreater action by girl --\textgreater boy\end{tabular}
\end{tabular}
}
    \label{tab:CIU_sequence_example}
\end{table}

\section{Spatio-Semantic Graphs and Features}\label{spatiosemantic-graph-features}
\label{sec:format}
A spatio-semantic graph can be constructed by mapping the CIUs to their respective locations, representing the visual path the speaker took when describing the picture. 
Each CIU is assigned a 2-dimensional coordinate on a pixel scale, with the Cookie Theft image sized at 546 $\times$ 290 pixels. 
The procedure for creating the spatio-semantic graph follows the implementation outlined in \cite{ambadi2021spatio}. 
The sequence of extracted CIUs is automatically encoded with their coordinate pairs. 
Using the NetworkX toolkit, a set of graph nodes and edges is constructed from these coordinates and their orderings. 
The resulting graph visualizes the nodes as the 23 CIUs in the picture, with the nodes representing the 23 CIUs in the picture and the edges representing the order of CIUs mentioned by the speaker. The distance between two connected CIUs is measured by the Euclidean distance. Additionally, each CIU can be assigned to one of the four quadrants of the picture. A graph can be constructed from the sequences of quadrant labels, where the nodes represent the quadrants and the edges illustrate how the speaker's attention transitions between different quadrants during the description of the picture. 

Once the graph is constructed based on the CIUs, a set of spatio-semantic features can be derived from the graph representation and the spatial position of the CIUs in the Cookie Theft picture. Table \ref{tab:spatio_semantic_features} defines the 12 spatio-semantic features computed in this study. These features are designed to capture the semantic order and summarize visuospatial relationships among the mentioned CIUs. 
They are able to describe aspects such as dispersal, efficiency, fixation, memory lapses and sporadicity in speech \cite{ambadi2021spatio}. 

\begin{table}[th!]
    \caption{List of spatio-semantic features and their definition.}
    \centering
        \resizebox{0.90\linewidth}{!}{
\begin{tabular}{l|l}
\textbf{Spatio-Semantic Features} & \textbf{Calculation}                                                                             \\ \hline
Avg. X                            & CIUs' mean X-coordinate                                                                          \\ \hline
Std. X                            & CIUs' standard deviation of X-coordinate                                                         \\ \hline
Avg. Y                            & CIUs' mean Y-coordinate                                                                          \\ \hline
Std. Y                            & CIUs' standard deviation of Y-coordinate                                                         \\ \hline
Total path distance               & Sum of all edge lengths in graph                                                                 \\ \hline
Unique nodes                      & CIUs count without duplicates                                                                    \\ \hline
Total path / Unique nodes        & \begin{tabular}[c]{@{}l@{}}Total path distance divided by \\ number of unique nodes\end{tabular} \\ \hline
Nodes                             & CIUs count With duplicates                                                                       \\ \hline
Self cycles                       & Count of consecutive same CIU                                                                    \\ \hline
Cycles                            & Count of repeated CIUs                                                                           \\ \hline
Self cycles (quadrants)           & Count of consecutive same quardrant                                                              \\ \hline
Cross ratio (quadrants)           & Ratio of inter-quadrant to intra-quadrant edges                                                 
\end{tabular}
}
    \label{tab:spatio_semantic_features}
\end{table}

\begin{table*}[th!]
    \caption{ANCOVA test results. (* $p<0.05$, ** $p<0.01$, *** $p<0.001$, $\dag$: Unique Nodes is used as the dependent variable, not as a covariate.)}
    \centering
        \resizebox{0.88\linewidth}{!}{
\begin{tabular}{l|cll|lll}
                        & \multicolumn{3}{c|}{\textbf{Using Manually Extracted CIUs}}                                                                                                                                                                                                           & \multicolumn{3}{c}{\textbf{Using Automatically Extracted CIUs}}                                                                                                                                                                                                                                   \\ \cline{2-7} 
                        & F-value & \multicolumn{1}{c}{\begin{tabular}[c]{@{}c@{}}Cognitively Unimpaired\\ Marginal Mean  (95\% CI)\end{tabular}} & \multicolumn{1}{c|}{\begin{tabular}[c]{@{}c@{}}Cognitively Impaired\\ Marginal Mean  (95\% CI)\end{tabular}} & \multicolumn{1}{c}{F-value} & \multicolumn{1}{c}{\begin{tabular}[c]{@{}c@{}}Cognitively Unimpaired\\ Marginal Mean  (95\% CI)\end{tabular}} & \multicolumn{1}{c}{\begin{tabular}[c]{@{}c@{}}Cognitively Impaired\\ Marginal Mean  (95\% CI)\end{tabular}} \\ \hline
Avg. X                  & 2.81          & 276 (274, 278)                                                                                                           & 271 (266, 276)                                                                                                          & 0.398                           & 263 (262, 265)                                                                                                           & 262 (258, 266)                                                                                                         \\
Std. X                  & 0.696            & 129 (128, 130)                                                                                                           & 130 (128, 131)                                                                                                          & 0.207                                & 127 (126, 127)                                                                                                     & 127 (126, 128)                                                                                                   \\
Avg. Y                  & 0.0000            & 173 (172, 174)                                                                                                           & 173 (171, 175)                                                                                                          & 0.264                                & 163 (162, 163)                                                                                                     & 162 (160, 164)                                                                                                   \\
Std. Y                  & 0.374           & 70.3 (69.8, 70.8)                                                                                                        & 69.9 (68.7, 71.1)                                                                                                       & 0.006                                & 70.5 (70.1. 70.9)                                                                                                        & 70.5 (69.7, 71.4)                                                                                                      \\
Total path distance     & 19.6***        & 1759 (1730, 1789)                                                                                                        & 1936 (1866, 2006)                                                                                                       & 39.5***                            & 2521 (2471, 2571)                                                                                                        & 2952 (2833, 3071)                                                                                                      \\
Total path/Unique nodes & 32.9***         & 136 (133, 138)                                                                                                           & 156 (150, 162)                                                                                                          & 37.5***                             & 164 (160, 167)                                                                                                           & 191 (183, 199)                                                                                                         \\
Self cycles            & 3.84           & 0.460 (0.418, 0.515)                                                                                                     & 0.594 (0.481, 0.708)                                                                                                    & 4.34**                              & 0.562 (0.506, 0.619)                                                                                                     & 0.723 (0.589, 0.857)                                                                                                   \\
Cycles                  & 30.73***        & 3.06 (2.89, 3.22)                                                                                                        & 4.30 (3.91, 4.70)                                                                                                       & 45.62***                            & 6.65 (6.37, 6.93)                                                                                                        & 9.23 (8.57, 9.90)                                                                                                      \\
Nodes                   & 29.0***        & 15.6 (15.5, 15.8)                                                                                                        & 16.9 (16.5, 17.4)                                                                                                       & 47.6***                            & 22.8 (22.5, 23.2)                                                                                                        & 26.2 (25.4, 27.1)                                                                                                      \\
Self cycles (Quadrants) & 2.20            & 6.72 (6.58, 6.86)                                                                                                        & 7.00 (6.67, 7.33)                                                                                                       & 39.3***                            & 10.8 (10.6, 11.0)                                                                                                         & 12.6 (12.1, 13.1)                                                                                                    \\
Cross ratio (quadrants) & 3.57          & 1.33 (1.28, 1.38)                                                                                                        & 1.46 (1.34, 1.58)                                                                                                       & 0.881                                & 1.11 (1.08, 1.14)                                                                                                        & 1.07 (1.00, 1.14)                                                                                                      \\
Unique nodes$\dag$           & 32.9***        & 13.0 (12.8, 13.2)                                                                                                        & 11.4 (10.9, 11.9)                                                                                                       & 46.4***                            & 15.8 (15.6, 16.0)                                                                                                        & 14.0 (13.6, 14.5)                                                                                                     
\end{tabular}
}
    \label{tab:ANCOVA_results}
\end{table*}

\section{Experimental Setup}
The spatio-semantic features derived from automatically extracted CIUs should produce similar results and conclusions as those obtained from manually extracted CIUs. 
In \cite{ambadi2021spatio}, statistical analyses were conducted to confirm group-level differences in spatio-semantic features between different clinical groups. We repeat the same analyses here for the automatic spatio-semantic graph. To automatically encode CIUs from a given Cookie Theft description, all punctuation in the original CHAT transcriptions is first removed. The text is then lemmatized using the SpaCy toolkit \cite{Honnibal_spaCy_Industrial-strength_Natural_2020}. 
The lemmatized text is then mapped to a sequence of CIUs using the dictionary in Table \ref{tab:CIU}. 
The spatio-semantic graph and its derived features are computed using the methods described in Section \ref{spatiosemantic-graph-features}.

To ensure a fair comparison between the manual and automated approaches for deriving spatio-semantic features, we followed the statistical analyses outlined in \cite{ambadi2021spatio}, which utilize analysis of covariance (ANCOVA). 
In this approach, each spatio-semantic feature is treated as the dependent variable in its own ANCOVA model. Demographic factors such as age, education, gender, and the count of unique nodes are included as covariates. The rationale for adjusting for unique nodes is based on the observation that the number of unique CIUs in the transcribed speech may differ between cognitively unimpaired and cognitively impaired groups; furthermore, this feature may correlate with all other features derived from the graph \cite{ambadi2021spatio}. 
In the ANCOVA test, a significance level of $p=0.05$ is used. 
A statistically significant difference between groups suggests that the feature distribution varies between the two diagnostic groups. 
The $F$-value, calculated as the ratio of between-group variance to within-group variance, is used to assess the extent of these differences, with larger $F$-values indicating more pronounced distinctions.

\begin{figure}[h!]
\centering

\subfigure[]{
\includegraphics[width=0.63\linewidth]{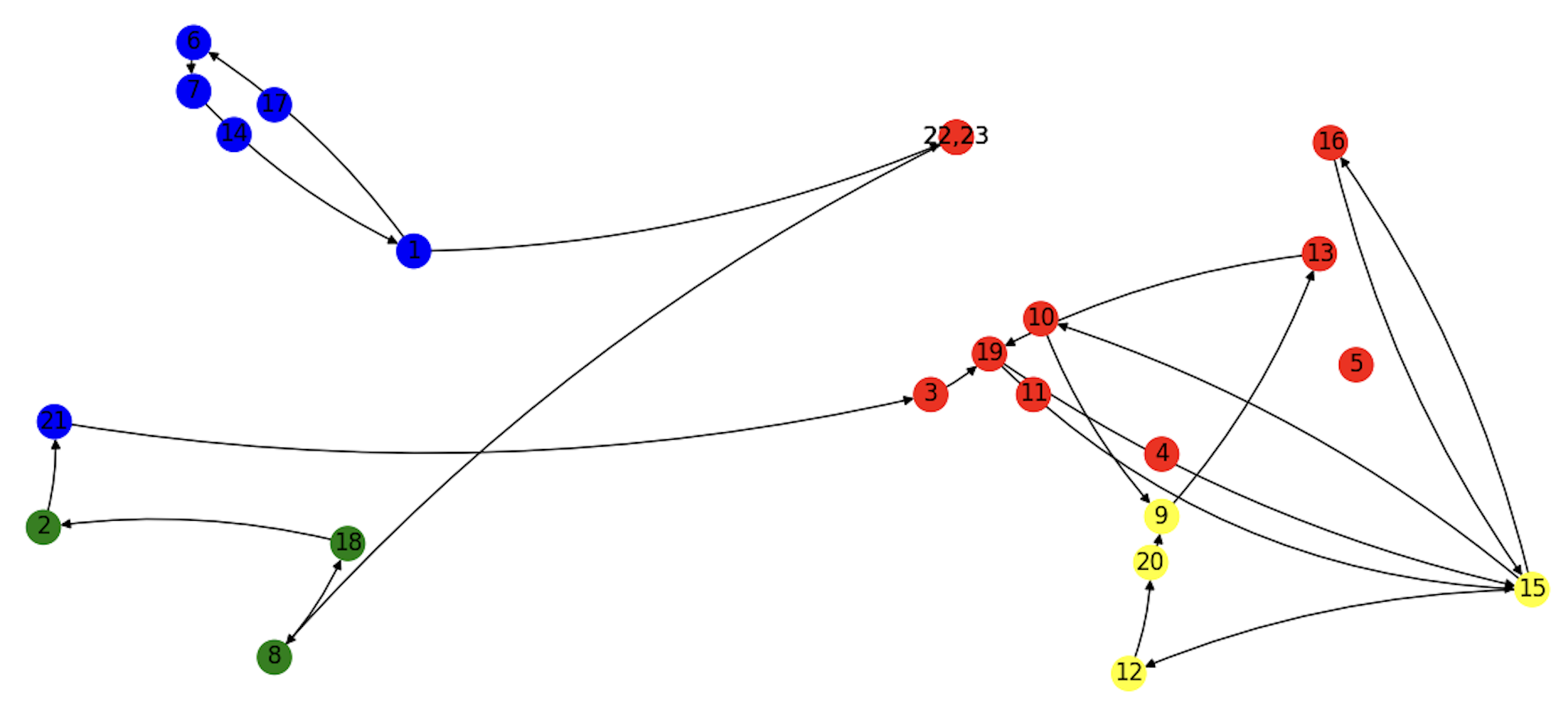}
}
\subfigure[]{
\includegraphics[width=0.63\linewidth]{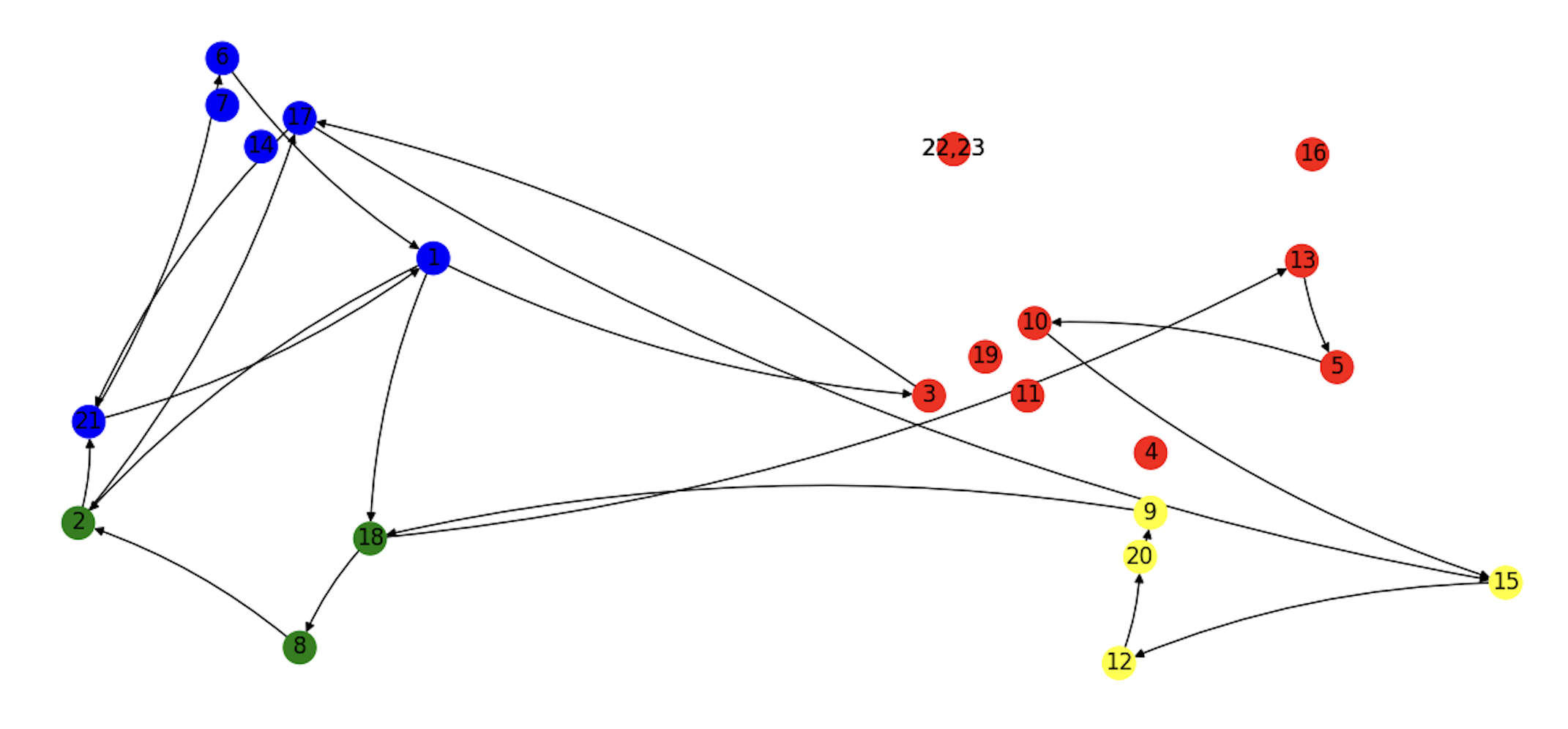}
}
\caption{Spatio-semantic graph derived from a) Cognitively-unimpaired speaker, b) Cognitively-impaired speaker.}
\label{fig:graph_visual}
\end{figure}

\section{Experimental Results And Discussion} 
Figure \ref{fig:graph_visual} illustrates examples of spatio-semantic graphs generated from automatically extracted CIUs for both cognitively unimpaired and cognitively impaired speakers. The graph for the cognitively impaired speaker diagnosed with dementia reveals an inefficient visual path characterized by frequent cross-quadrant transitions and a longer total path. 
In contrast, the unimpaired speaker’s path is more organized, with fewer repetitions and shorter total distances. These spatio-semantic features derived from the graphs are expected to reflect the differences in cognitive function between the two groups. 

Table \ref{tab:ANCOVA_results} presents the ANCOVA analysis results of spatio-semantic features derived from both manually and automatically extracted CIUs, along with the 95\% confidence intervals for each feature. The total path distance computed from automatically extracted CIUs is generally longer than that from manually extraction. This difference is consistent with the earlier illustration in Table \ref{tab:CIU_sequence_example}, where the automated method captured more CIUs. 
Most features show statistical significance, regardless of the CIU extraction method used. 
When comparing manual and automatic spatio-semantic graphs, differences were observed only in the self cycles (according to both CIUs and quadrants) derived from automatically extracted CIUs. 
Since the automated approach detects more CIUs, there is a higher likelihood of consecutive CIUs occurring within the same quadrant, as indicated by the confidence interval. This altered distribution of CIUs across quadrants could lead to slightly different conclusions in the ANCOVA test.

For all spatio-semantic features with significant group differences,  those derived from automatically extracted CIUs yield higher $F$-values than those from manual extraction, indicating greater distinction between groups. Overall, the ANCOVA results suggest that the proposed dictionary in Table \ref{tab:CIU} effectively supports the automatic extraction of CIUs and spatio-semantic features, achieving performance comparable to or better than the manual method. 
Considering the importance of automatic dementia detection in clinical speech analytics, these findings imply that researchers can extract spatio-semantic features without human annotation of CIUs and use them as inputs for clinical speech models. Additionally, these features are linked to clinically relevant constructs, offering  clinical interpretability.

\section{Conclusion}
This study proposes an approach for automatically extracting content information units (CIUs) from the Cookie Theft picture description task. A dictionary containing words relevant to the CIUs was constructed, enabling the creation of spatio-semantic graphs. These graphs allow the derivation of spatio-semantic features that can distinguish cognitively impaired speakers from cognitively unimpaired ones. Experimental results demonstrate that the automated extraction method accurately identifies more CIUs than the manual approach, offering a more detailed description of the speakers’ visual path during the task. ANCOVA analyses of the spatio-semantic features indicate that the automated method produces conclusions similar to the manual approach while yielding greater group differences across each feature. 
Future work will explore using large language models for CIU extraction, without the need for manual dictionary design.

\bibliographystyle{IEEEtran} 
\bibliography{strings,refs}

\end{document}